\documentclass[letterpaper, 10 pt, conference]{ieeeconf}  

\IEEEoverridecommandlockouts                              

\usepackage{graphics} 
\usepackage{epsfig} 
\usepackage{amsmath} 
\usepackage{amssymb}  
\usepackage{graphicx}
\usepackage{cite}
\usepackage{soul}
\usepackage{environ}
\usepackage{multirow}
\usepackage[table]{xcolor}
\usepackage{url}
\usepackage{color}

\usepackage{xcolor}
\usepackage{mathtools}
\usepackage{siunitx}
\usepackage{adjustbox}
\usepackage{float}
\usepackage[font=small,tableposition=top]{caption}
\usepackage{booktabs}

\usepackage{pgfplots}
\pgfplotsset{compat=newest} 
\usetikzlibrary{plotmarks}
\usepgfplotslibrary{patchplots}
\usepackage{grffile}

\usepackage{gensymb}

\NewEnviron{scaletikzpicturetowidth}[1]{%
        \def\tikz@width{#1}%
        \begin{lrbox}{\measure@tikzpicture}%
        \BODY
        \end{lrbox}%
        \pgfmathparse{#1/\wd\measure@tikzpicture}%
        \BODY
}
\makeatother

\graphicspath{{./}{./pics/}{../pics}}
\DeclareGraphicsExtensions{.jpg,.JPG,.png,.PNG,.pdf,.PDF,.eps,.EPS}

\usepackage[hidelinks]{hyperref}

\title{\LARGE \bf
Self-supervised 6D Object Pose Estimation for Robot Manipulation
}

\author{Xinke Deng$^{1, 2}$, Yu Xiang$^{1}$, Arsalan Mousavian$^{1}$, Clemens Eppner$^{1}$, Timothy Bretl$^{2}$, Dieter Fox$^{1, 3}$
\thanks{* This work is done while the first author was an intern at NVIDIA. $^{1}$NVIDIA, $^{2}$University of Illinois at Urbana-Champaign, $^{3}$University of Washington}%
}

\begin{document}

\maketitle
\thispagestyle{empty}
\pagestyle{empty}

\begin{abstract}

To teach robots skills, it is crucial to obtain data with supervision. Since annotating real world data is time-consuming and expensive, enabling robots to learn in a self-supervised way is important. In this work, we introduce a robot system for self-supervised 6D object pose estimation. Starting from modules trained in simulation, our system is able to label real world images with accurate 6D object poses for self-supervised learning. In addition, the robot interacts with objects in the environment to change the object configuration by grasping or pushing objects. In this way, our system is able to continuously collect data and improve its pose estimation modules. We show that the self-supervised learning improves object segmentation and 6D pose estimation performance, and consequently enables the system to grasp objects more reliably. A video showing the experiments can be found at {\color{blue}\href{https://youtu.be/W1Y0Mmh1Gd8}{https://youtu.be/W1Y0Mmh1Gd8}}.

\end{abstract}


\section{INTRODUCTION}

For robots to gain skills such as manipulation, learning-based methods have received more attention recently due to their capability in handling the complexity in robot perception. For instance, a number of 6D object pose estimation methods using deep neural networks are introduced for model-based grasping \cite{xiang2017posecnn,wang2019densefusion}. Learning techniques also demonstrate impressive performance in grasping unknown objects \cite{mahler2017dex,levine2018learning,mousavian20196}. It is well known that training deep neural networks requires a significant amount of data. How to obtain training data becomes the bottleneck in applying learning techniques to different robotic problems these days.

It is very appealing to conduct training in simulation, since synthetic data generated from simulators are free, come with ground truth annotations, and are easy to scale up. However, due to the domain gap, models trained exclusively with synthetic data cannot be guaranteed to work well in the real world. In contrast, training can be performed using data directly collected in the real world. However, annotating real world data is time consuming and labor intensive.

Recently, the concept of self-supervised learning is becoming attractive, where robots autonomously annotate real world data for training. Self-supervised learning has great potential to improve the robustness of robotic systems and achieve life-long learning for robots. The main challenge in self-supervision is how to annotate the real world data automatically, and meanwhile, guarantee the accuracy of the annotations. Previous works have explored self-supervised learning for model-free grasping~\cite{levine2018learning}, pixel-wise object segmentation~\cite{zeng2017multi}, and object detection with bounding boxes~\cite{mitash2017self}. However, to the best of our knowledge, no work has been proposed for directly annotating 6D object pose in a self-supervised fashion. This is mainly because accurately estimating 6D object pose and automatically evaluating the estimation is challenging. 

\begin{figure}
    \centering
    \includegraphics[width=0.45\textwidth]{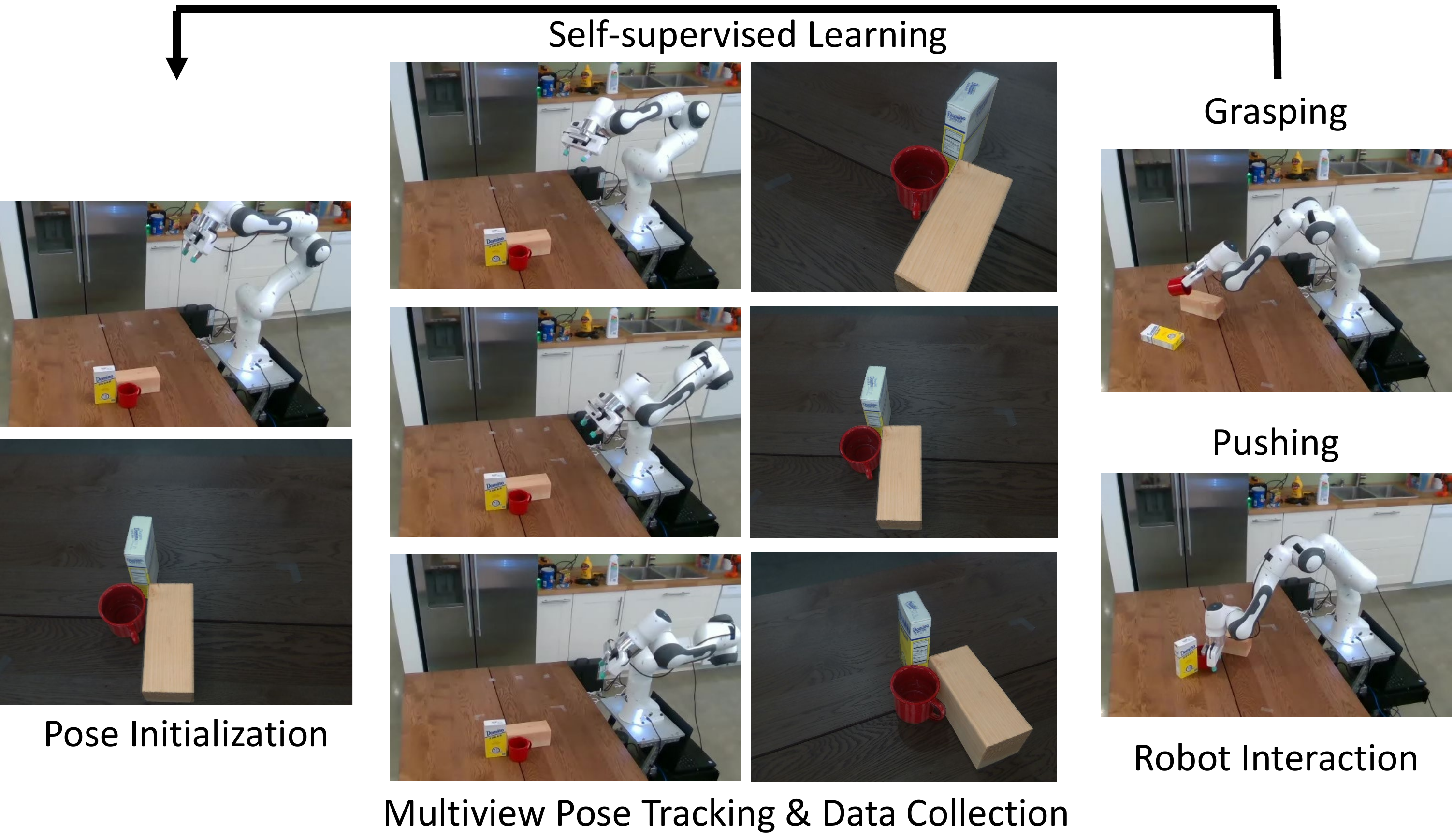}
    \caption{Our self-supervised learning process for robot manipulation.}
    \label{fig:intro}
    \vspace{-5mm}
\end{figure}

In this work, by leveraging the recent advances in 6D object pose estimation with deep neural networks~\cite{xiang2017posecnn,sundermeyer2018implicit,deng2019poserbpf}, we present a novel self-supervised 6D object pose estimation system for manipulation~(Fig.~\ref{fig:intro}). An active vision system is set up by mounting an RGB-D camera onto the hand of a robot manipulator. Given a set of objects in the workspace, we control the robot arm to collect images from different views. Meanwhile, the 6D poses of these objects are accurately initialized and then 
tracked based on the PoseRBPF framework~\cite{deng2019poserbpf}, where we use the forward kinematics of the robot arm as a motion prior in the particle filtering. Consequently, our system is able to collect images of multiple objects from different views with accurate 6D pose annotations. We can use this data to perform self-supervised training of the deep neural networks in the system for object segmentation and pose estimation. After each iteration of the multi-view pose tracking, the robot manipulator grasps or pushes one object in the scene to generate a new scene, and then repeats the pose tracking process. In this way, our system is able to continuously generate new data for self-supervised learning, and improve its pose estimation and grasping performance in a life-long learning fashion.

\begin{figure*}
    \centering
    \includegraphics[width=0.78\textwidth]{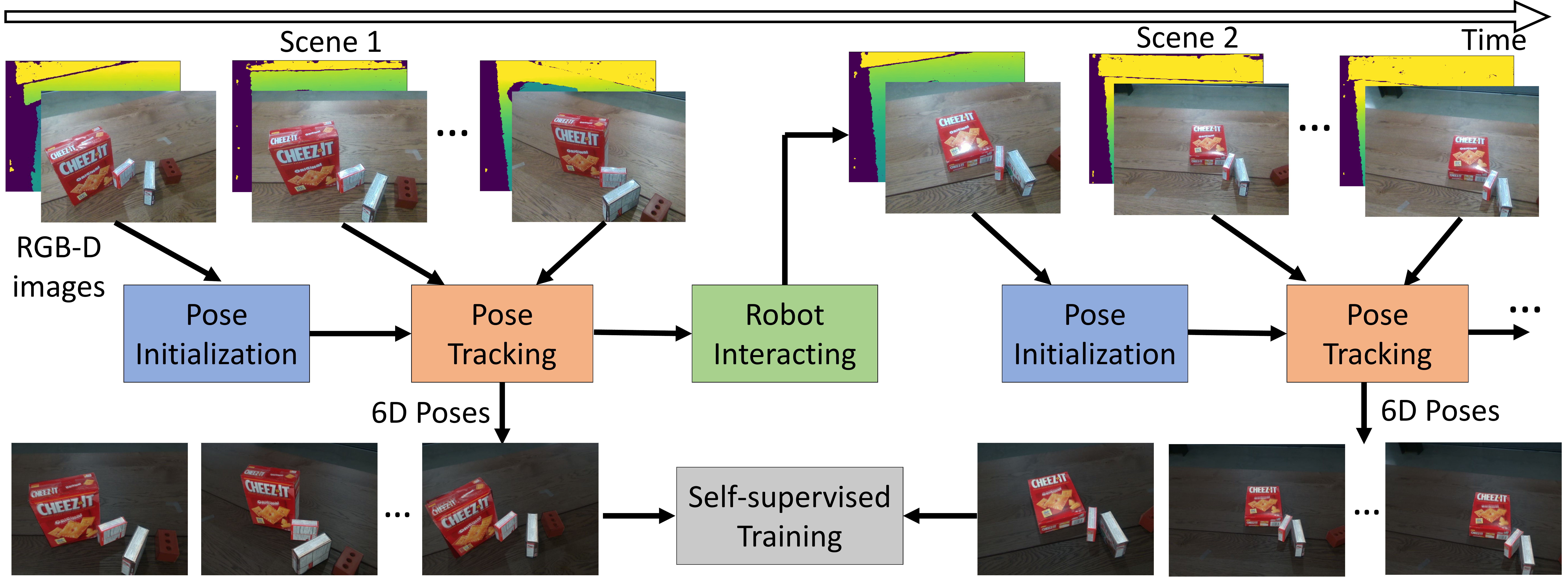}
    \caption{Overview of our self-supervised 6D object pose estimation system.}
    \label{fig:system}
    \vspace{-5mm}
\end{figure*}

We conduct experiments to demonstrate how the self-supervised learning improves the system. We first show that by using the self-annotated images to fine-tune the semantic segmentation network, we can significantly improve the segmentation accuracy compared to the original network trained with synthetic data only. Semantic segmentation suffers from the domain gap severely for objects with complex material properties, such as shininess. Fine-tuning the network can bridge the domain gap. Second, we show that fine-tuning the auto-encoders used in PoseRBPF~\cite{deng2019poserbpf} can improve their reconstruction quality, and consequently improve the pose estimation accuracy. Finally, we conduct robot grasping experiments to illustrate that our fine-tuned system can achieve a high success rate in model-based grasping based on 6D object pose estimation.

This paper is organized as follows: After discussing related
work, we introduce our system for self-supervised 6D object pose estimation, followed by experiments and a conclusion.

\section{RELATED WORK}

\textbf{6D Object Pose Estimation.} Given the 3D models of objects, the goal of 6D pose estimation is to estimate the 3D translation and the 3D rotation of these objects from input images. Traditionally, the problem is tackled by local feature matching \cite{lowe1999object,rothganger20063d,collet2011moped} or template matching \cite{hinterstoisser2011gradient,hinterstoisser2012model,cao2016real}. Recently, learning-based methods receive more attention due to their ability to generalize to different backgrounds, lighting, and occlusions \cite{brachmann2014learning,bo2014learning,krull2015learning}. Deep neural networks further boost the performance by learning more discriminative features and classifiers \cite{xiang2017posecnn,kehl2017ssd,sundermeyer2018implicit,deng2019poserbpf,wang2019densefusion}. However, deep learning methods also require more data for training.

A few recent methods show that it is possible to train exclusively with synthetic data and apply the trained network to the real world \cite{tremblay2018deep,sundermeyer2018implicit}. But for objects with mismatch between simulation and real world, these methods tend to fail, such as mismatch in texture materials or lighting. Meanwhile, some efforts are devoted to collecting large-scale real-world dataset for 6D pose estimation \cite{xiang2017posecnn,marion2018label}. These methods use a moving camera around a static scene so that the pose labels for a video sequence can be generated with labeling one frame and tracking the camera ego-motion \cite{xiang2017posecnn}  or labeling the 3D reconstructed scene \cite{marion2018label}. Although tracking camera motion simplifies the labeling process, these methods require human to arrange the scene, collect video sequences of objects, and initialize the systems, which is difficult to scale to a large number of scenes. In this work, we employ robots to automatically arrange scenes and generate accurate 6D pose annotations without pose initialization from human.

\textbf{Self-supervised Learning.} Due to its ability to learn from unlabeled data, self-supervised learning has been studied in different sub-fields in AI, such as in robotics~\cite{levine2018learning, agrawal2016learning}, computer vision~\cite{doersch2015unsupervised,larsson2016learning}, machine learning~\cite{raina2007self} and natural language processing~\cite{devlin2018bert}. The majority of these methods focus on learning from fixed datasets. In robotics, since the robot can interact with the real world, we can plan and control how data is collected ~\cite{finn2017deep, pathak2018learning}. This degree of freedom in data collection significantly boosts the self-supervised learning efficiency.

Previous works on self-supervised learning in robot perception mainly focus on learning object segmentation~\cite{zeng2017multi}, object detection~\cite{mitash2017self}, and dense pixel-wise correspondences~\cite{schmidt2016self,florence2018dense}, since it is relatively easy to obtain annotations in 2D image space.
In this work, we explore self-supervised learning for 6D object pose estimation. Our system benefits from the recent advances in 6D object pose estimation using deep neural networks~\cite{xiang2017posecnn,sundermeyer2018implicit,deng2019poserbpf}. By incorporating these techniques into the interactive robot perception framework~\cite{bohg2017interactive}, we can obtain accurate 6D poses of objects to improve the system.

\section{SELF-SUPERVISED 6D OBJECT POSE ESTIMATION SYSTEM}


\begin{figure*}
    \centering
    \includegraphics[width=0.87\textwidth]{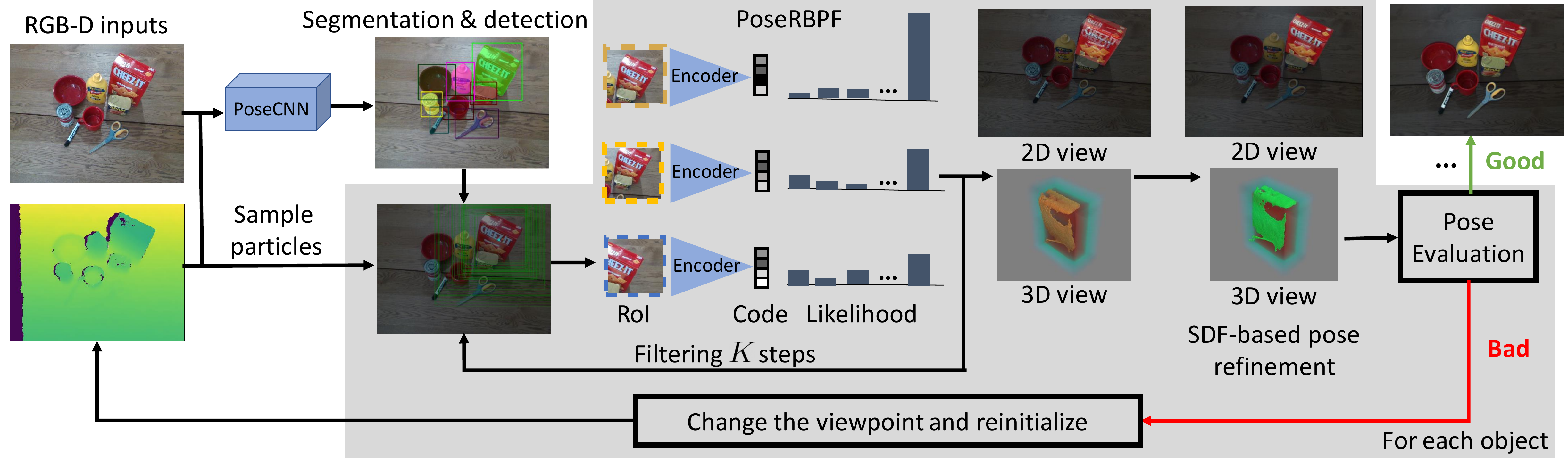}
    \caption{Pose initialization process. This module takes an RGB-D image as input and estimates the 6D poses of all the objects in the image. The process iterates until all object poses are accurate according to the pose evaluation component.}
    \label{fig:pose_init}
    \vspace{-5mm}
\end{figure*}

\subsection{System Overview}

In our system, a RGB-D camera is mounted to the hand of a robot manipulator, where images from different views can be collected by moving the robot arm~(Fig.~\ref{fig:intro}). The system consists of a pose initialization module, a pose tracking module, a robot interaction module and a self-supervised training module~(Fig.~\ref{fig:system}). 1) Given a set of objects in the robot's workspace, the pose initialization module estimates the 6D poses of all objects from a single RGB-D image. To ensure accurate pose initialization, we evaluate the estimated poses and re-initialize the unsatisfied objects until all the poses are accurate. 2) Then the pose tracking module tracks the 6D poses of these objects when the robot arm moves to different predefined way points. This step also collects images with 6D object poses from different views, and saves the data in an online fashion. Both the pose initialization and pose tracking are based on the PoseRBPF framework using particle filtering~\cite{deng2019poserbpf}. 3) After the data is saved, the robot interaction module controls the manipulator to either grasp or push one object in the scene, thereby generating a new scene. In this way, we can continuously collect data with different arrangements of objects. 4) The self-supervised training module fine-tunes the deep neural networks in the system with the saved data to boost its performance.


\subsection{The Pose Initialization Module}

Given a set of objects in a scene, the task of the pose initialization module is to accurately estimate the 6D poses of all objects from a single RGB-D image. Based on this estimate we can start pose tracking~(Fig.~\ref{fig:pose_init}). Pose estimation from a single image is a challenging problem. With a new scene, the camera is moved automatically to a viewpoint with high elevation angle, so the pose estimation is less affected by the occlusions between objects.

\subsubsection{Object Segmentation and Detection}

The first step is to detect the objects in the input image. In general, any state-of-the-art object detector can be used, such as YOLO~\cite{redmon2016you}, SSD~\cite{liu2016ssd} or Faster R-CNN~\cite{ren2015faster}. Instead of using these top-down methods that are built on bounding boxes, we resort to PoseCNN~\cite{xiang2017posecnn}. PoseCNN first labels every pixel into an object class and then applies Hough voting to find the centers and bounding boxes of the objects. PoseCNN generates segmentation masks of the objects which can be used to segment the point clouds of the objects in pose refinement~\cite{wong2017segicp}. In addition, the object centers from PoseCNN are more accurate than bounding box centers from the top-down methods thanks to the Hough voting strategy.



\subsubsection{3D Rotation and 3D Translation Estimation}

To estimate the 3D rotation $\mathbf{R}$ and 3D translation $\mathbf{t}$ of objects in the scene, we utilize the PoseRBPF framework~\cite{deng2019poserbpf}. PoseRBPF is a Rao-Blackwellized particle filter combined with a learned auto-encoder network~\cite{sundermeyer2018implicit} for 6D object pose estimation. In PoseRBPF, each particle consists of a 3D translation and a full distribution over~$\mathbf{SO}(3)$. A 3D translation can be represented by $(u, v, z)$, where $(u, v)$ is the object center in the image and $z$ is the distance of object center from the camera in 3D.

For each detected object, the 3D translations of the particles in PoseRBPF are sampled according to the detected object center $(\hat{u}, \hat{v})$ and the observed depth image~$\mathbf{D}$: $(u_i, v_i) \sim \mathcal{N}\big((\hat{u}, \hat{v}), \sigma_{u}, \sigma_{v}\big)$, $z_i \sim \mathcal{U}\big(\mathbf{D}(u_i, v_i)-\frac{d}{2}, \mathbf{D}(u_i, v_i)+\frac{d}{2})$, where $(u_i, v_i, z_i)$ denotes the translation of the $i$th particle, $\sigma_{u}, \sigma_{v}$ denote the standard deviations of the Gaussian distributions for object center, and $d$ denotes the range of a uniform distribution where $z_i$ is sampled from. $\mathbf{D}(u_i, v_i)$ is the depth at pixel location $(u_i, v_i)$. Each particle determines a Region of Interest (RoI) of the target object~(Fig.~\ref{fig:pose_init}). The RoI is fed into the auto-encoder to compute a code which is used to measure the likelihood of the particle in PoseRBPF. After the initial sampling of the particles, the system performs $K$ filtering steps with the same RGB-D image to ensure the convergence of particles, from which we can extract the expectation of 3D rotation and 3D translation.


\subsubsection{6D Pose Refinement}
\label{sec:sdf}
Since the number of particles in PoseRBPF is limited due to time budget and discretization of 3D rotation in each particle, we perform a continuous optimization using 3D points from the depth image to refine the poses.
Finding correct pixels corresponding to the object is crucial in the refinement process. We estimate the segmentation mask of the object $\mathbf{\Omega}$ by considering both the predicted segmentation mask $\mathbf{\hat{\Omega}}$ from PoseCNN and the predicted 6D pose $(\bar{\mathbf{t}}, \bar{\mathbf{R}})$ from PoseRBPF. We first render the object according to the estimated pose. Denoting the rendered depth image and segmentation mask of the object as $\bar{\mathbf{D}}$ and $\bar{\mathbf{\Omega}}$. The final segmentation mask $\mathbf{\Omega}_{\text{obj}}$ is estimated with $\mathbf{\Omega}_{\text{obj}} = \hat{\mathbf{\Omega}} \cap \bar{\mathbf{\Omega}} \cap \mathbf{\Omega}_{\mathbf{D}}$. Here, $\mathbf{\Omega}_{\mathbf{D}}$ represents the pixels where the rendered depth matches with the measured depth within a margin $m$: $\mathbf{\Omega}_{\mathbf{D}}=\left \{  \forall (u, v), \left |  \mathbf{D}(u, v) - \mathbf{\bar{D}}(u, v)\right | < m\right \}$. Then the point cloud of the object $\mathbf{P}_{\text{obj}}$ can be computed by back-projecting the pixels in $\mathbf{\Omega}_\mathbf{D}$:
\begin{equation} \label{eq:points}
    \mathbf{P}_{\text{obj}}=\left \{ \mathbf{D}(u, v)\mathbf{K}^{-1}(u, v, 1)^{T}, (u, v)\in \mathbf{\Omega}_{obj} \right \},
\end{equation}
where $\mathbf{K}$ represents the intrinsic matrix of the camera.

After computing the 3D points on the object, we optimize the pose by matching these points against the Signed Distance Function~(SDF) of the object model as in~\cite{schmidt2014dart}. The optimization problem we solve is
\begin{equation}
    (\hat{\mathbf{t}}, \hat{\mathbf{R}})= \underset{\mathbf{t}, \mathbf{R}}{\arg\min} \sum_{\mathbf{p}_i \in \mathbf{P}_{\text{obj}}}\left | \mathbf{SDF}_{\text{obj}}(\mathbf{p}_i, \mathbf{t}, \mathbf{R}) \right | + \lambda \frac{1}{2} \| \mathbf{t} - \bar{\mathbf{t}} \| ^2,
\end{equation}
where $\mathbf{p}_i$ is a 3D point in the point cloud~$\mathbf{P}_{\text{obj}}$, $\mathbf{SDF}_{\text{obj}}(\mathbf{p}_i, \mathbf{t}, \mathbf{R})$ denotes the signed distance value by transforming the point $\mathbf{p}_i$ from the camera coordinate into the object model coordinate using pose $(\mathbf{t}, \mathbf{R})$, and $\lambda$ is a weight to balance the regularization term on the translation. Fig.~\ref{fig:pose_init} illustrates the matching between the point cloud against the SDF before and after the optimization.



\subsubsection{Pose Evaluation}
\label{sec:evaluation}
\begin{figure}
    \centering
    \includegraphics[width=0.42\textwidth]{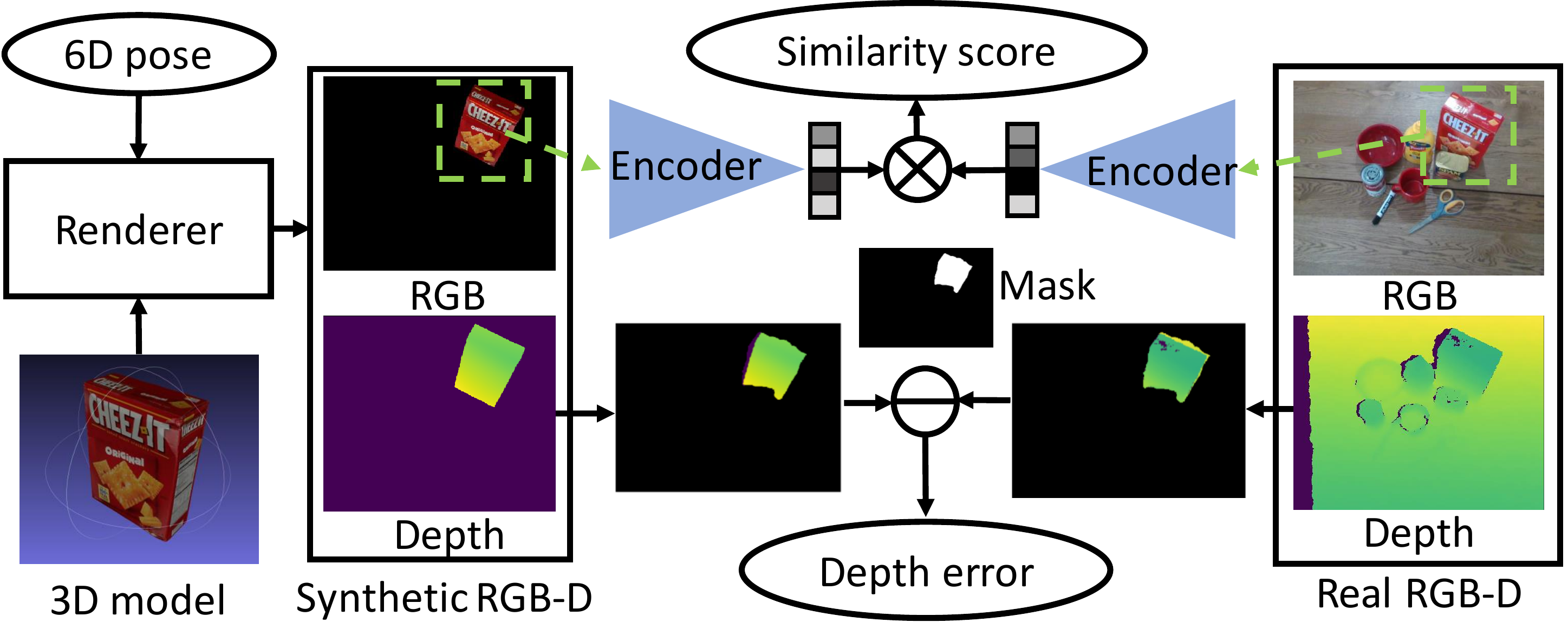}
    \caption{The pose evaluation component compares the rendered RGB-D image with the real RGB-D image.}
    \label{fig:pose_evaluate}
    \vspace{-5mm}
\end{figure}

Due to noise in both RGB and depth measurements, failure in pose estimation is inevitable. To detect failures and label object poses accurately, we propose a pose evaluation process~(Fig.~\ref{fig:pose_evaluate}) so that objects with erroneous pose estimation will be reinitialized. We first render an object according to its estimated pose to generate a synthetic RGB-D image. Then we compute two metrics by comparing the synthetic image with the real image. For RGB, we compute the code of the color rendering~$\hat{\mathbf{c}}$ and the code of the color measurement~${\mathbf{c}}$ by passing the two corresponding RoIs through the auto-encoder in PoseRBPF. The cosine distance $s = (\hat{\mathbf{c}}\cdot\mathbf{c})/(\left \|\hat{\mathbf{c}} \right \|\cdot \left \| \mathbf{c}\right \|)$ between the two codes is then computed to indicate the pose confidence in the color space. For depth, we compare the rendered depth image $\hat{\mathbf{D}}$ and the measured depth image ${\mathbf{D}}$ using the mask~$\mathbf{\Omega}_{\text{obj}}$ from Eq.~\eqref{eq:points}. The mean depth error is computed to measure the pose accuracy in the depth space: $e = \sum_{(u, v) \in \mathbf{\Omega}_{\text{obj}}} |\hat{\mathbf{D}}(u, v) - \mathbf{D}(u, v)|$.
Finally, we declare pose estimation failure if $s < s^*$ or $e > e^*$, where $s^*$ and $e^*$ are predefined thresholds. In this case the system re-initializes the pose of the failed object.

\subsection{The Pose Tracking Module}


Once object poses in the scene are initialized, the robot moves the camera to different views around the objects for data collection. Meanwhile, a continuous sequence of video frames is collected in real time, and the poses of the objects in the sequence are tracked with the PoseRBPF framework. The pose tracking module generates accurate pose annotations online for self-supervised training and enables the robot to interact with the objects continuously.

In order to track all objects accurately, we first propagate all particles according to the robot's forward kinematics. Denoting the camera pose transformation from the forward kinematics as $(\Delta\mathbf{t}, \Delta\mathbf{R}) \in \mathbf{SE}(3)$, the particles are propagated according to the following motion prior:
\begin{align}
    P(\mathbf{t}_k|\mathbf{t}_{k-1}, \Delta\mathbf{t}, \Delta\mathbf{R}) &= \mathcal{N}(\Delta\mathbf{R}\mathbf{t}_{k-1} + \Delta\mathbf{t}, \Sigma_\mathbf{t}), \\
    P(\mathbf{R}_k|\mathbf{R}_{k-1}, \Delta\mathbf{t}, \Delta\mathbf{R}) &= \mathcal{N}(\Delta\mathbf{R}\mathbf{R}_{k-1}, \Sigma_\mathbf{R}),
\end{align}
where $\Sigma_\mathbf{t}$ and $\Sigma_\mathbf{R}$ denote the covariance matrix of the Gaussian distributions of the translation and the rotation, respectively.
Since the rotation in PoseRBPF is represented as a \emph{full discrete distribution} over $\mathbf{SO}(3)$, the distributions need to be shifted entirely. We first represent $\Delta\mathbf{R}$ as $(\Delta\phi, \Delta\theta, \Delta\psi)$ according to the discretization of the rotation distribution, where $\Delta\phi$, $\Delta\theta$, $\Delta\psi$ represent the changes in pitch, yaw and roll, respectively. Then the rotation distribution is adjusted by performing bilinear interpolation of the shifted grids $\left \{ \phi+\Delta\phi, \theta+\Delta\theta, \psi+\Delta\psi\right \}$.


Due to the noises from the joint encoders of the robot or the bias in the hand-eye calibration, simply applying forward kinematics might often lead to inaccurate object poses. Similar to the pose initialization process, we perform~\SI{1} step filtering with the propagated particles using the current RGB-D image for all the objects. To ensure the quality of the generated data for self-supervised learning, we only save data when the two pose evaluation metrics are above certain thresholds~(Sec.~\ref{sec:evaluation}). Once the system decides to save the data, the object poses are further refined with the SDF-based pose refinement algorithm as described in Sec.~\ref{sec:sdf}.


\begin{figure}
    \centering
    \includegraphics[width=0.38\textwidth]{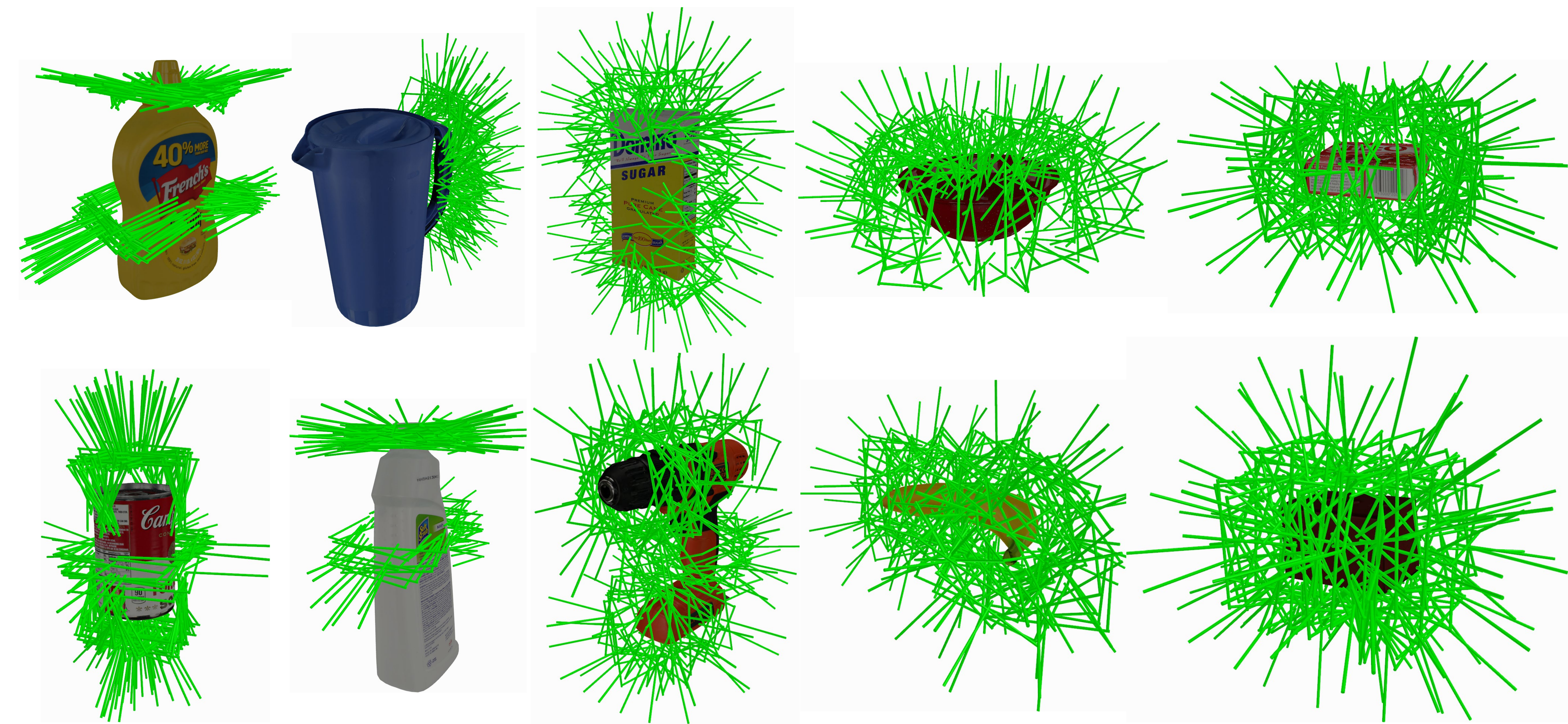}
    \caption{Examples of 100~precomputed grasps for 10 YCB objects.}
    \label{fig:grasps}
    \vspace{-5mm}
\end{figure}

\subsection{The Robot Interaction Module}

After a scene is captured, the robot physically interacts with objects in the scene to change their configuration. This way the system can continuously capture new data with different poses of the objects. Two types of interactions are used. The robot either pushes an object in a random direction or grasps an object and places it at a different location~(Fig.~\ref{fig:intro}).

Pushing is defined by choosing a direction and random radius with respect to the object center and executing the motion. To ensure that objects stay in the workspace of the robot, objects are pushed either toward the center of mass of other objects in the clutter or the center of the robot workspace. For grasping, a set of diverse grasps for each object is computed offline, where we use 100 parallel-jaw grasps sampled from the robust grasps of~\cite{EppnerISRR2019}~(see Fig.~\ref{fig:grasps} for examples). During online execution, using the estimated pose of the object, the pre-computed grasps are transformed from object to camera coordinates. Then the robot selects any grasp that is kinematically feasible and collision-free with other objects using the RRT$^*$ planning algorithm in MoveIt!~\cite{chitta2012moveit}. After grasping the object, the robot first rotates its gripper to change the orientation of the object. Then a random placement location is chosen within the robot workspace and the robot places the object.

\subsection{The Self-Supervised Training Module}

With the collected images and estimated object poses, we can fine-tune the neural networks in our system to improve the pose estimation performance. Initially, both the network for segmentation and detection in PoseCNN and the auto-encoders used in PoseRBPF~(see Fig.~\ref{fig:pose_init}) are trained only with synthetic data. As a result, the system cannot even segment some textureless objects in our experiments such as a red foam brick. Therefore, we employ the curriculum learning idea~\cite{bengio2009curriculum} to bootstrap the system sequentially. We start with simple scenes that contain only one object. If the network cannot segment the object, we perform a global initialization in PoseRBPF, where particles are uniformly sampled in the image. This process collects training data with single objects in images. We can then fine-tune the segmentation network to make it segment all objects. Fine-tuning with single objects also enables the network to segment objects in cluttered scenes. In the next stage, we move to cluttered scenes with multiple objects, and collect training data with different object poses and occlusions to fine-tune the segmentation network and the auto-encoders. During fine-tuning, we mix synthetic and real data to prevent the networks from overfitting. We can iterate between data collection and fine-tuning. Because when the networks become better, the system is more efficient in collecting new data.

\section{EXPERIMENTS}

\begin{table} \setlength{\tabcolsep}{4pt}
    \centering
    \caption{F1 score of semantic segmentation for 20 YCB objects on the test set. The best F1 score for each object are bold, and numbers below 60\% are highlighted red.}
    \begin{tabular}{|c|c|c|c|c|c|c|}
\hline
Model                    & Synth.                      & \begin{tabular}[c]{@{}c@{}}+20\%\\ Real\end{tabular} & \begin{tabular}[c]{@{}c@{}}+40\%\\ Real\end{tabular} & \begin{tabular}[c]{@{}c@{}}+60\%\\ Real\end{tabular} & \begin{tabular}[c]{@{}c@{}}+80\%\\ Real\end{tabular} & \begin{tabular}[c]{@{}c@{}}+100\%\\ Real\end{tabular} \\ \hline
master\_chef\_can   & 69.3                        & 88.7                                                  & 92.8                                                  & 91.9                                                  & 91.6                                                  & \textbf{93.9}                                          \\
cracker\_box        & 84.7                        & 92.8                                                  & 93.0                                                  & \textbf{93.4}                                         & 93.0                                                  & \textbf{93.4}                                          \\
sugar\_box          & 83.0                        & 92.0                                                  & 92.0                                                  & 92.4                                                  & 92.5                                                  & \textbf{92.6}                                          \\
tomato\_soup\_can   & 83.6                        & 90.2                                                  & 90.5                                                  & 90.8                                                  & 91.2                                                  & \textbf{91.4}                                          \\
mustard\_bottle     & 83.9                        & 92.5                                                  & 93.3                                                  & 93.3                                                  & 93.9                                                  & \textbf{94.2}                                          \\
tuna\_fish\_can     & {\color[HTML]{FE0000} 42.3} & 90.1                                                  & 90.2                                                  & \textbf{91.7}                                         & \textbf{91.7}                                         & 91.6                                                   \\
pudding\_box        & 61.6                        & 85.7                                                  & 84.6                                                  & 85.9                                                  & \textbf{87.1}                                         & 87.0                                                   \\
gelatin\_box        & 66.6                        & 83.6                                                  & 83.2                                                  & 83.7                                                  & 82.0                                                  & \textbf{84.6}                                          \\
potted\_meat\_can   & 62.9                        & 84.1                                                  & 85.4                                                  & 86.4                                                  & 86.6                                                  & \textbf{88.6}                                          \\
banana              & 79.8                        & 87.3                                                  & 88.2                                                  & 89.0                                                  & 89.0                                                  & \textbf{89.3}                                          \\
pitcher\_base       & {\color[HTML]{FE0000} 51.5} & 86.3                                                  & 84.3                                                  & 88.0                                                  & \textbf{89.7}                                         & 89.6                                                   \\
bleach\_cleanser    & {\color[HTML]{FE0000} 57.9} & 89.3                                                  & 92.1                                                  & 93.3                                                  & 90.2                                                  & \textbf{93.4}                                          \\
bowl                & 69.8                        & 90.4                                                  & 92.5                                                  & 93.2                                                  & 94.5                                                  & \textbf{95.4}                                          \\
mug                 & 69.2                        & 90.3                                                  & 90.9                                                  & 91.4                                                  & \textbf{92.0}                                         & 91.0                                                   \\
power\_drill        & 66.1                        & 84.4                                                  & 87.3                                                  & 88.0                                                  & 87.5                                                  & \textbf{88.5}                                          \\
wood\_block         & 64.2                        & 82.6                                                  & 80.2                                                  & 85.1                                                  & 86.0                                                  & 86.1                                                   \\
scissors            & {\color[HTML]{FE0000} 36.3} & 71.9                                                  & 75.8                                                  & 77.4                                                  & 77.5                                                  & \textbf{78.9}                                          \\
large\_marker       & {\color[HTML]{FE0000} 55.5} & 73.8                                                  & 75.6                                                  & 76.2                                                  & 75.4                                                  & \textbf{77.1}                                          \\
extra\_large\_clamp & {\color[HTML]{FE0000} 15.5} & 76.3                                                  & 76.0                                                  & \textbf{79.1}                                         & 77.1                                                  & \textbf{79.1}                                          \\
foam\_brick         & {\color[HTML]{FE0000} 12.2} & 86.5                                                  & 86.7                                                  & 87.6                                                  & 86.6                                                  & \textbf{88.9}                                          \\ \hline
MEAN                     & 60.8                        & 85.9                                                  & 86.7                                                  & 87.9                                                  & 87.8                                                  & \textbf{88.7}                                          \\ \hline
\end{tabular}
\label{tab:ycb_seg} 
\vspace{-4mm}
\end{table}

\subsection{Data Collection}

We conduct experiments with 20~objects in the YCB Object and Model Set~\cite{calli2015ycb}. During around 12 robot hours, our system collected 497 scenes, 6,541 RGB-D images and 22,851 object instances in these images with accurate 6D poses. In this process, human intervention is required in two cases: 1) whenever a new subset of the 20~YCB objects needs to be presented to the robot, and 2) to rearrange objects if the system cannot initialize the object poses for a while. Compared to human collected videos for pose annotation, our system saves a significant amount of human labor, and has the ability to scale up to a large number of scenes. For comparison, the YCB-Video dataset~\cite{xiang2017posecnn} and the LabelFusion dataset~\cite{marion2018label} have 92 scenes and 138 scenes, respectively. We split the collected data into a training set (265 scenes, 3,590 images) and a test set (232 scenes, 2,951 images).

\begin{figure}
    \centering
    \includegraphics[width=0.42\textwidth]{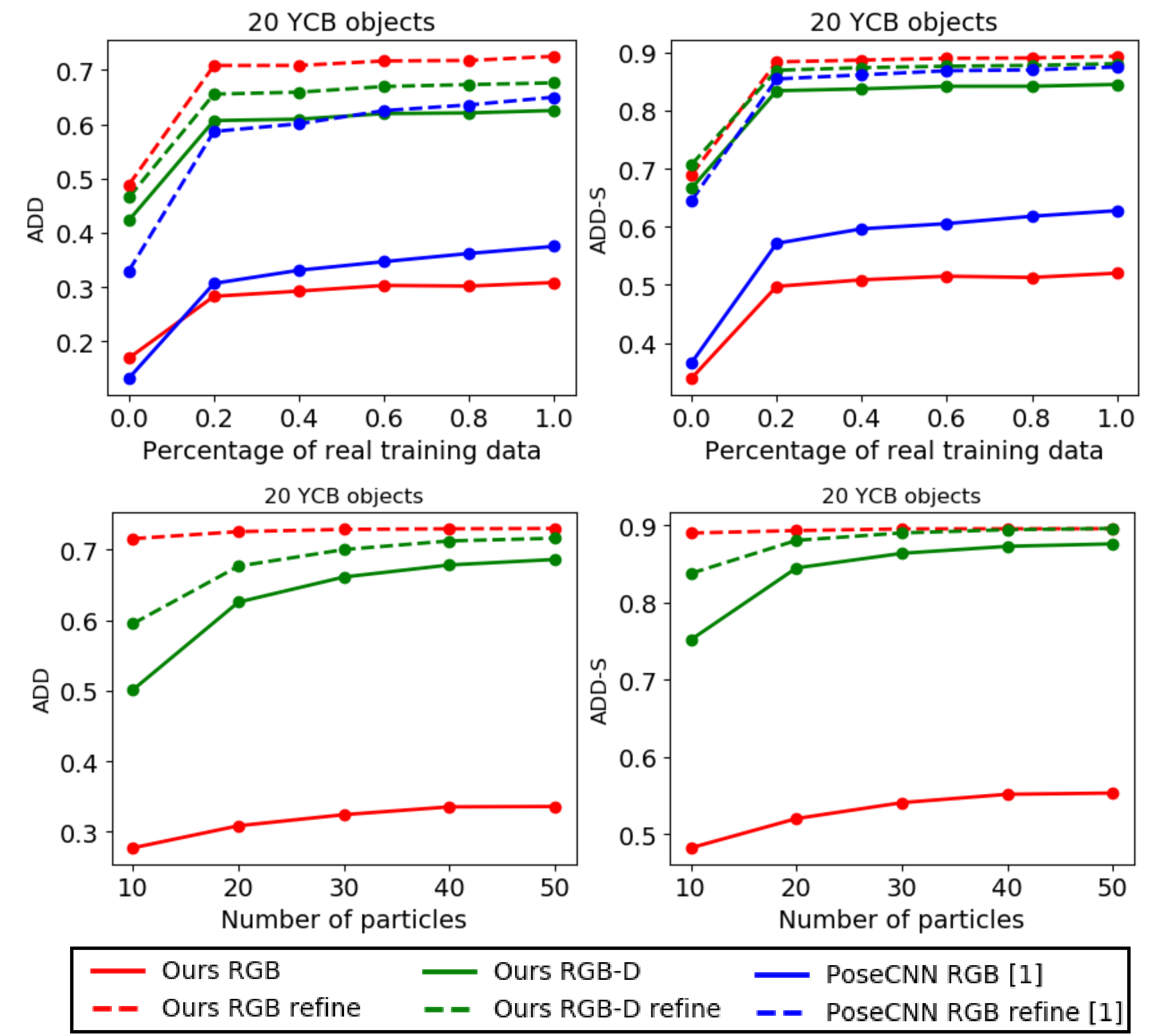}
    \caption{Evaluation of 6D object pose estimation on the test set.}
    \label{fig:pose_comparison}
    \vspace{-4mm}
\end{figure}

\subsection{Evaluation Metrics}

We evaluate how the system improves by fine-tuning the neural networks with the collected data. For semantic segmentation, we compute the F1 score by comparing the predicted labels with the ground truth labels pixel-wisely. For evaluating 6D poses, we use ADD and ADD-S metric as defined in~\cite{hinterstoisser2012model,xiang2017posecnn}:
$\textsc{ADD} = \frac{1}{m}\sum_{\mathbf{x} \in \mathcal{M}}\| (\mathbf{R} \mathbf{x} + \mathbf{t}) - (\mathbf{\tilde{R}} \mathbf{x} + \mathbf{\tilde{t}})  \|$,
$\textsc{ADD-S} = \frac{1}{m}\sum_{\mathbf{x}_1 \in \mathcal{M}} \min_{\mathbf{x}_2 \in \mathcal{M}} \| (\mathbf{R} \mathbf{x}_1 + \mathbf{t}) - (\mathbf{\tilde{R}} \mathbf{x}_2 + \mathbf{\tilde{t}})  \|$,
where $\mathcal{M}$ denotes the set of 3D model points and $m$ is the number of points. $(\mathbf{R}, \mathbf{t})$ and $(\mathbf{\tilde{R}}, \mathbf{\tilde{t}})$ are the ground truth pose and estimated pose, respectively. To evaluate the grasping performance, we compute the success rate, and the speed for initializing pose estimations as well as grasping.

\subsection{Implementation Details}
We implement our system on a 7-DoF Franka Panda manipulator with an Intel RealSense D415 camera mounted on its parallel-jaw gripper. The deep neural networks run on two NVIDIA Titan Xp GPUs. In the pose initialization module, the noise for sampling particles $\sigma_u$, $\sigma_v$, and $\sigma_z$ are set to \SI{20}{pixels}, \SI{20}{pixels}, and \SI{0.1}{m} respectively. In pose refinement, the margin for computing the visibility mask is set to \SI{0.02}{m}, and the weight $\lambda$ is set to \SI{0.001}. In pose evaluation, thresholds for deciding estimation failures are \SI{0.5} and \SI{0.03}{m} for $s^*$ and $e^*$. In pose tracking module, process noise $\Sigma_\mathbf{t}$ and $\Sigma_\mathbf{R}$ are set to \SI{0.015}{m} and \SI{0.05}{rad}.



\subsection{Semantic Segmentation}

By fine-tuning the segmentation network with self-collected images, we can bridge the domain gap and significantly improve the segmentation accuracy. Table~\ref{tab:ycb_seg} shows the F1 scores of 20~YCB objects on the test set. The performance of the networks fine-tuned with different percentages of real training data is presented. By using \SI{20}{\percent} training data, the overall F1 score is increased by \SI{25}{\percent}. Fine-tuning with more data consistently improves the performance. In Table~\ref{tab:ycb_seg}, we highlight objects on which synthetic training performs poorly. Most of them are textureless (pitcher\_base, wood\_block, extra\_large\_clamp and foam\_brick). Objects with metal parts generate reflections which are not well modeled in simulation (tuna\_fish\_can, potted\_meat\_can and scissors). For pudding\_box and power\_drill, the textures of the real objects are different from the 3D models due to product updates. large\_marker is difficult since it is small in images. All these issues make synthetic training difficult to transfer to the real world, and our self-supervised learning system successfully adapts to the new environment.

\subsection{6D Object Pose Estimation}

In this experiment, we evaluate 6D object pose estimation using single RGB-D images as input. If the system can estimate object poses quickly and accurately from just a single image, it would significantly speed up manipulation tasks. We evaluate our pose initialization method~(Fig.~\ref{fig:pose_init}) and the original PoseCNN~\cite{xiang2017posecnn}. Both methods use the same network for segmentation and detection. The main difference is in rotation and translation estimation. PoseCNN~\cite{xiang2017posecnn} is trained to directly regress to object distance and 3D rotation, while our method uses the auto-encoders and particle filter to estimate translation and rotation.

\begin{figure}
    \centering
    \includegraphics[width=0.46\textwidth]{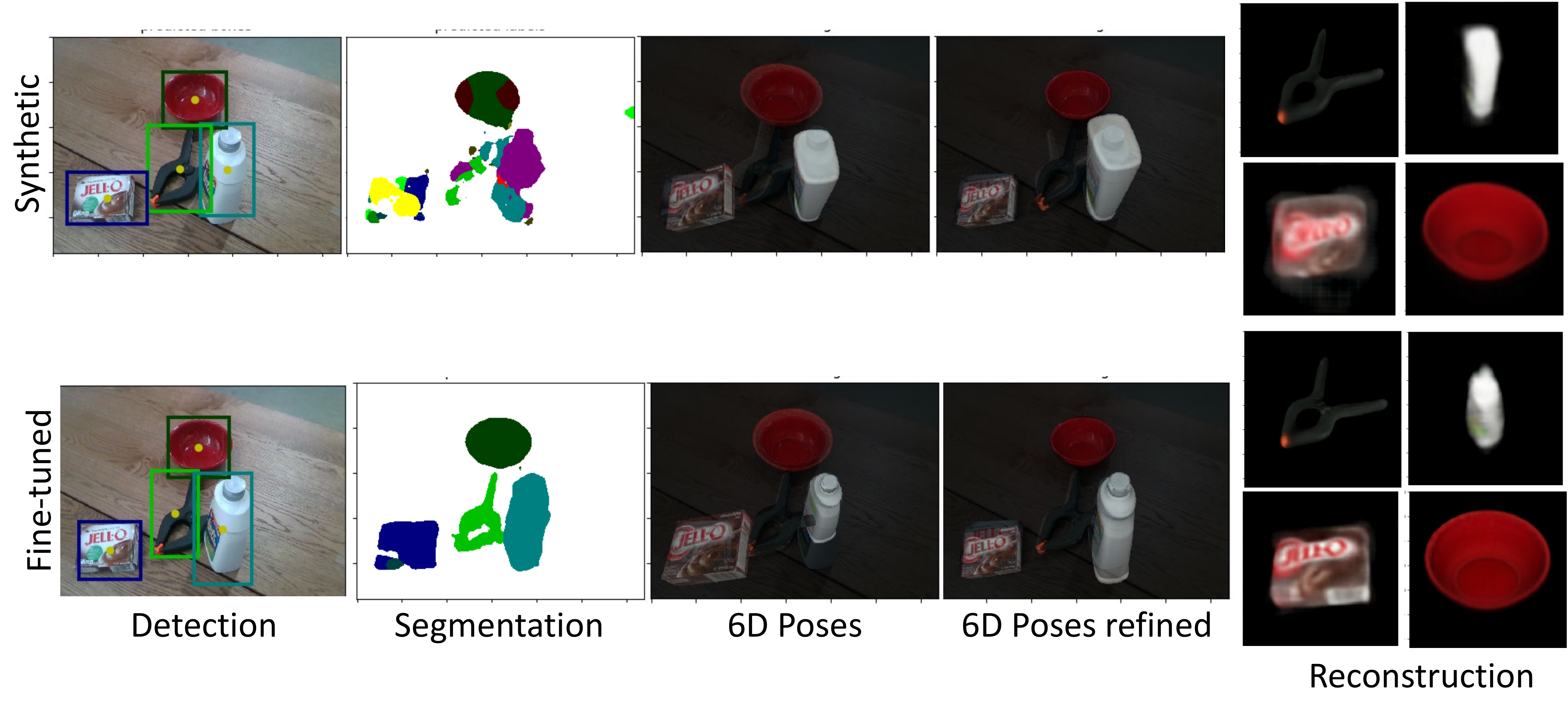}
    \caption{An example where synthetic networks fail in segmentation and pose estimation, but the fine-tuned networks succeed.}
    \label{fig:anecdotal}
    \vspace{-4mm}
\end{figure}

Fig.~\ref{fig:pose_comparison} presents the 6D pose estimation accuracy in terms of ADD and ADD-S using different percentages of real images for fine-tuning or number of particles. We can see that 1) both methods improve consistently with more real training data, and the gap between synthetic models and fine-tuned ones is obvious. 2) Using depth is important to achieve high accuracy, and our SDF-based pose refinement algorithm significantly boosts the accuracy. 3) The performance of our method improves consistently with more particles in PoseRBPF. However the improvement becomes less obvious when performing refinement. 4) It is interesting that our RGB based particle filter with SDF refinement achieves the best performance, which is even better than the RGB-D particle filter with refinement. This is because the auto-encoders are trained only with RGB images. Using RGB images in the particle filter obtains more accurate rotation, and then SDF refinement can fix errors in translation, and adjust rotation locally. Fig~\ref{fig:anecdotal} shows an example on how the fine-tuned segmentation network and autoencoders improve segmentation and pose estimation.

\begin{figure}
    \centering
    \includegraphics[width=0.42\textwidth]{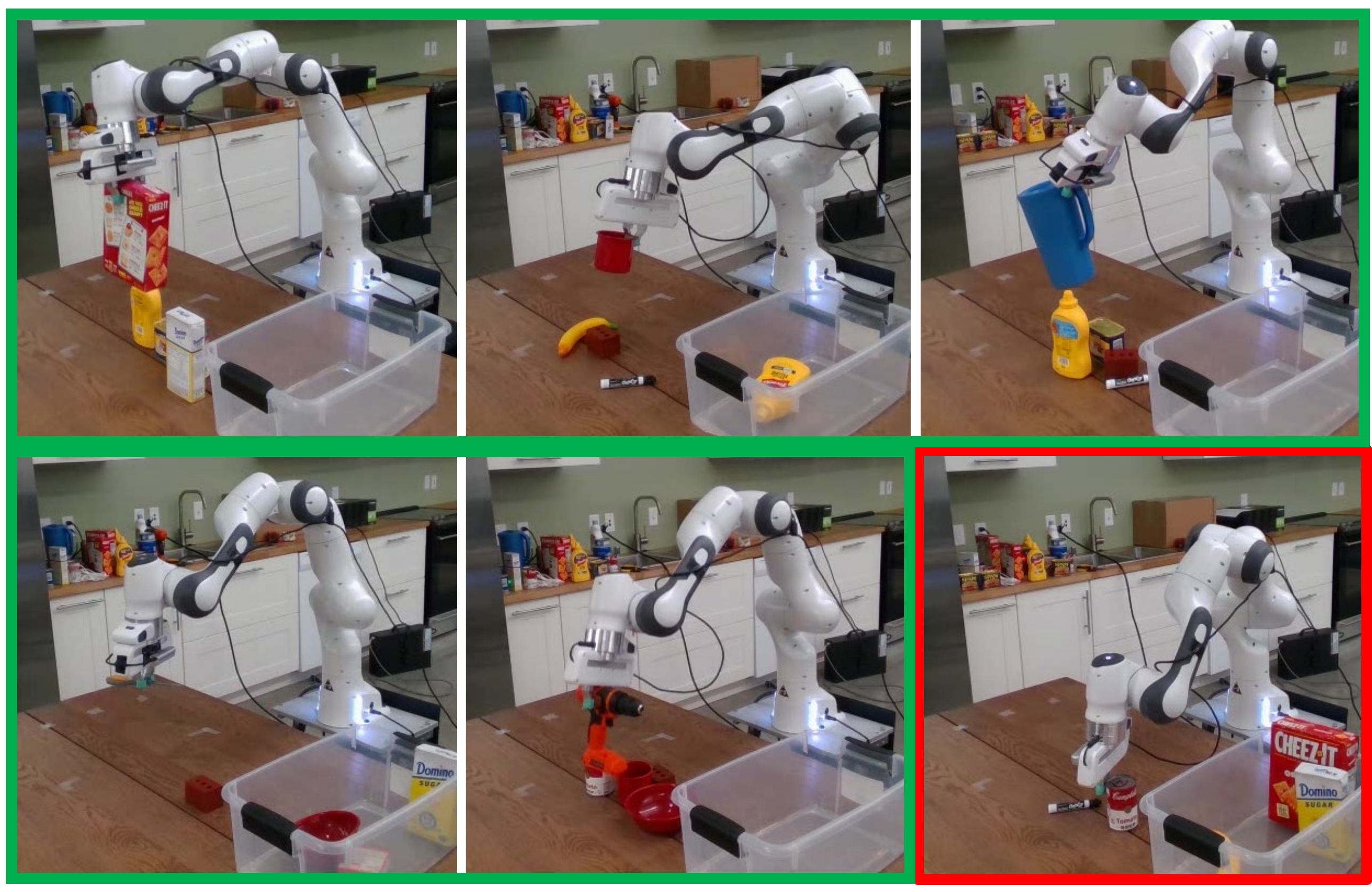}
    \caption{Grasping performed in real cluttered environments. Here we show 5 successful and 1 failed trials on grasping YCB objects.}
    \label{fig:grasp experiments}
    \vspace{-1mm}
\end{figure}

\subsection{Real Robot Grasping Experiments}

We evaluate the performance of our system on a pick-and-place task. The goal is to pick up all objects in the scene and place them into a bin. The distance between camera and clutter is around~\SI{0.5}{m} and the initial elevation angle of the camera is~$55^\circ$. The system first initializes all objects in the scene. Every time before grasping the target object, all objects are evaluated and those with a detected failure are re-initialized based on the target object. The estimated poses are used for grasping the objects. After picking up an object, the robot follows a predefined trajectory to place it. 

We created six different cluttered environments. Every environment consists of five different YCB objects. In total, 14~graspable YCB objects are selected, and 30~grasping trials are executed. Grasping examples are shown in Fig.~\ref{fig:grasp experiments}. We evaluate three metrics: 1) grasp success rate, 2) average time elapsed for accurately initializing each object, and 3) average time elapsed for initializing and grasping each object. The results are shown in Table~\ref{tab:grasping}. It is possible that the pose of the objects cannot be successfully initialized. We allow~\SI{45}{s} for initializing each object before calling timeout and labelling the trial as a failure. As we see, by fine-tuning with additional self-annotated data, the system's performance on grasping can be significantly improved: the success rate is improved by~\SI{85.7}{\percent}, the duration for initialization is reduced by~\SI{55.3}{\percent}, the overall duration for grasping an object is reduced by~\SI{27.3}{\percent}. There were four cases when grasps failed. Three of them were due to errors in pose estimations; in the other one MoveIt! failed to generate a feasible trajectory.

\begin{table}[]
\centering
\caption{Statistics for 30 grasping trials in clutter.}
\begin{tabular}{|c|c|c|c|}
\hline
                                                                            Grasp Trials & \begin{tabular}[c]{@{}c@{}}Success \\rate [\%]\end{tabular} & \begin{tabular}[c]{@{}c@{}}Avg. initialization \\ time per object [s]\end{tabular} & \begin{tabular}[c]{@{}c@{}}Avg. grasp\\duration [s]\end{tabular}\\ \hline
synthetic data only                                                          & 46.70                                                                 & 10.04 (std: 15.98)
     & 21.27 \\ \hline
\begin{tabular}[c]{@{}c@{}}synthetic + real data\end{tabular} & \textbf{86.70}                                                        & \textbf{4.48} (std: 5.71)
                  & \textbf{15.47} \\ \hline
\end{tabular}

\label{tab:grasping}
\vspace{-6mm}
\end{table}

\section{CONCLUSION}

We introduce a novel self-supervised 6D object pose estimation system for robot manipulation. Our system is able to automatically collect real world images with accurate 6D object poses to fine-tune its neural networks for pose estimation. In the learning stage, the robot interacts with objects in a scene by grasping or pushing them to create new scenes, so new data can be collected continuously. We show that, starting from neural networks trained with synthetic images only, our system consistently improves the pose estimation accuracy through self-supervised learning. The grasping success rate and grasping time are also significantly improved. For future work, we plan to investigate how to improve grasp selection and grasp execution with self-supervised learning in our system.


\bibliographystyle{IEEEtran}

\bibliography{references}

\end{document}